\documentclass[10pt,journal]{IEEEtran}

\usepackage[pdftex]{graphicx}
\DeclareGraphicsExtensions{.pdf,.jpeg,.png}

\usepackage{multirow,setspace,verbatim,amsfonts,graphicx,amsmath,amsthm,amsbsy,amssymb,epsfig}
\usepackage{url,cite}
\usepackage{booktabs}

\usepackage{array}
\usepackage{booktabs}

\begin{document}

	\title{A Novel Adaptive Kernel for the RBF Neural Networks}
	\author{Shujaat~Khan, Imran~Naseem, 
		Roberto~Togneri,~\IEEEmembership{Senior~Member,~IEEE}
		and~Mohammed~Bennamoun,~\IEEEmembership{Senior~Member,~IEEE} 
		\thanks{Shujaat Khan is with the Faculty of Engineering Science and Technology, Iqra University, Defence View, Shaheed-e-Millat Road (Ext.) 
			Karachi-75500, Pakistan.  Email: shujaat@iqra.edu.pk}
		\thanks{Imran Naseem and Roberto Togneri are with the School of Electrical, Electronic and Computer Engineering, The University of Western Australia, 35 Stirling Highway, Crawley, WA 6009, Australia.  Email: \{imran.naseem,roberto.togneri\}@uwa.edu.au}
		\thanks{Mohammed Bennamoun is with the School of Computer Science and Software Engineering, The University of Western Australia, 35 Stirling Highway, Crawley, WA 6009, Australia.  Email: mohammed.bennamoun@uwa.edu.au. }
		\thanks{This work was published in Circuits, Systems, and Signal Processing (CSSP) \cite{khan2017novel}.}}

	\maketitle
	
	\begin{abstract}
		In this paper, we propose a novel adaptive kernel for the radial basis function (RBF) neural networks. The proposed kernel adaptively fuses the Euclidean and cosine distance measures to exploit the reciprocating properties of the two.  The proposed framework dynamically adapts the weights of the participating kernels using the gradient descent method thereby alleviating the need for predetermined weights.  The proposed method is shown to outperform the manual fusion of the kernels on three major problems of estimation namely nonlinear system identification, pattern classification and function approximation.
	\end{abstract}
	
	\begin{IEEEkeywords}
	Artificial neural networks, Radial basis function, Gaussian kernel, Support Vector Machine, Euclidean distance, Cosine distance, Kernel fusion
	\end{IEEEkeywords}
	
	\IEEEpeerreviewmaketitle

	\section{Introduction}
	\label{intro}
	The RBF neural networks have shown excellent performance in a number of problems of practical interest. In \cite{New3} the reservoirs of brine are analyzed for physicochemical properties using the RBF neural networks with the genetic algorithms.  The proposed model is called the GA-RBF model and has shown good results compared to the previous approaches.  In \cite{New5} the RBF kernel is used to predict the pressure gradient with high accuracy.  In the context of nuclear physics, RBF has been effectively used to model the stopping power data of materials as in \cite{New7}.  A comprehensive discussion of various applications can be found in \cite{New10}.
	
	In the recent years, considerable advancement has been made in the field.  In \cite{New2} a couple of new RBF construction algorithms are proposed with the aim of increasing error convergence rates with fewer computational nodes.  The first method expands popular Incremental Extreme Learning Machine algorithms by adding Nelder-Mead simplex optimization. The second algorithm uses Levenberg-Marquardt algorithm to optimize the positions and heights of RBF.  The results have shown better error performance compared to the previous research.  A new architecture of the optimized RBF neural network classifier is developed with the aid of fuzzy clustering and data preprocessing techniques in \cite{New4}.  In \cite{New6} a bee-inspired algorithm, called cOptBees, has been used with heuristics to automatically select the number, location and dispersions of basis functions to be used in  the RBF networks. The resultant BeeRBF is shown to be competitive and has the advantage of automatically determining the number of centers.  To accelerate the learning for the large-scale data sequence an incremental learning algorithm is proposed in \cite{New8}.  The merits of fuzzy and crisp clustering are effectively combined in \cite{New9}.
	
	In \cite{New20} orthogonal least-square based alternative learning procedure is proposed. In the algorithm, the centers of the RBF are selected one by one in a rational way until an adequate network has been constructed. 
	In \cite{New11} a novel RBF network with the multi-kernel is proposed to obtain an optimized and flexible regression model. The unknown centres of the multi-kernels are determined by an improved k-means clustering algorithm. An orthogonal least squares (OLS) algorithm is used to determine the remaining parameters. 
	Another learning algorithm proposed in \cite{New19} simplifies the neural network training through the use of an adaptive computation algorithm (ACA). The convergence of the ACA is analyzed by the Lyapunov criterion. 
	In \cite{New18} a sequential framework Meta-Cognitive Radial Basis Function network (McRBFN) and its Projection based Learning (PBL) referred to as PBL-McRBFN is proposed. The PBL-McRBFN is inspired by human meta-cognitive learning principles. The proposed algorithm is evaluated on two practical problems namely, the acoustic emission signal classification and the mammogram for cancer classification. 
	In \cite{New16} a non-parametric supervised classifier based on neural networks is proposed and is referred to as Self Adaptive Growing Neural Network (SAGNN). The SAGNN allows a neural network to adapt its size and structure according to the training data.  The performance of the method is evaluated for fault diagnosis and compared with various non-parametric supervised neural networks. 
	A hybrid optimization strategy is proposed in \cite{New12} by incorporating the adaptive optimization of particle swarm optimization (PSO) into a genetic algorithm (GA), named the HPSOGA. The proposed strategy is used for determining the parameters of radial basis function neural networks automatically (e.g., the number of neurons and their respective centers and radii). 

	\begin{figure}[h!]
		\begin{center}
			\centering 
			\includegraphics*[scale=0.4,bb=50 0 600 350]{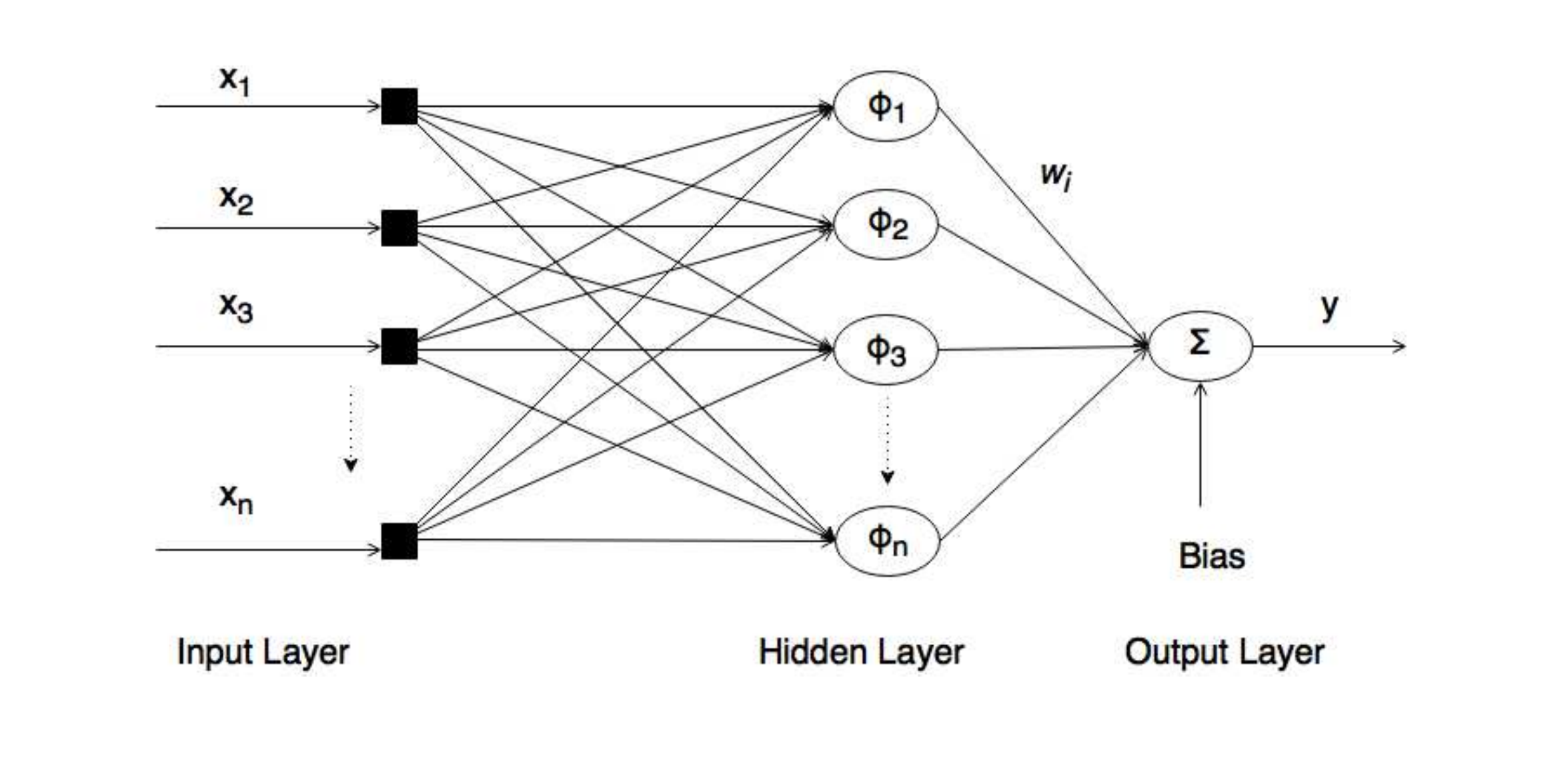} 
		\end{center}
		\caption{Architecture of an RBF neural network.}
		\label{rbfarch}
	\end{figure}
	
	Essentially the architecture of RBF networks consists of three layers:  (1) an input layer, (2) a nonlinear hidden layer, and (3) a linear output layer, refer to Figure \ref{rbfarch}.  Let $\mathbf{x} \in \mathbb{R}^{m_0}$ be the input vector, then the overall mapping of the RBF network, $s:\mathbb{R}^{m_0}\rightarrow\mathbb{R}^{1}$, is given as:
	
	\begin{eqnarray}
	y=\sum_{i=1}^{m_1}w_i\phi_i(\left\|\mathbf{x}-\mathbf{x}_i\right\|)+b
	\end{eqnarray} 
	
	where $m_1$ is the number of neurons in the hidden layer, $\mathbf{x}_i \in \mathbb{R}^{m_0}$ are the centers of the RBF network, $\mathbf{w}_i$ are the synaptic weights connecting the hidden layer to output neuron, $b$ is the bias term of the output neuron and $\phi_i$ is the basis function of the $i^{th}$ hidden neuron.  Without loss of generality and for simplicity a single output neuron is considered.  Conventional RBF networks employ a number of kernels such as multiquadrics, inverse multiquadrics and Gaussian (\cite{fifth}). The Gaussian kernel, due to its versatility, is considered to be the most commonly used kernel (\cite{seventeen}):
	
	\begin{eqnarray}
	\phi_i(\left\|\mathbf{x}-\mathbf{x}_i\right\|)=\exp\left(\frac{-\left\|\mathbf{x}-\mathbf{x}_i\right\|^2}{\sigma^{2}}\right)
	\end{eqnarray}   
	
	where $\sigma$ is the spread of the Gaussian kernel.  In one form or the other, the kernels use the concept of distance measure with the centers of the network.  Conventionally, the Euclidean distance has been used as an efficient distance metric.  Recently, it has been argued that the cosine distance metric has some complimentary properties to offer compared to the Euclidean distance measure (\cite{drmoin}):  
	\begin{eqnarray}
	\label{gamma1}
	\phi_{i1}(\mathbf{x}.\mathbf{x}_i)=\frac{\mathbf{x}.\mathbf{x}_i}{\left\|\mathbf{x}\right\|\left\|\mathbf{x}_i\right\| + \gamma}
	\end{eqnarray} 
	where the term $\gamma > 0$, a very small constant, is added to the denominator to avoid the indeterminant form of (\ref{gamma1}) in case $\left\|\mathbf{x}\right\|$ or $\left\|\mathbf{x}_i\right\|$ is zero.  Accordingly a novel kernel has been proposed to fuse the cosine and Euclidean distances (\cite{drmoin}):
	\begin{eqnarray}
	\label{novel}
	\phi_i(\mathbf{x},\mathbf{x}_i)=\alpha_1\phi_{i1}(\mathbf{x}.\mathbf{x}_i)+\alpha_2\phi_{i2}(\left\|\mathbf{x}-\mathbf{x}_i\right\|)
	\end{eqnarray}
	
	where $\phi_{i1}(\mathbf{x}.\mathbf{x}_i)$ and $\phi_{i2}(\left\|\mathbf{x}-\mathbf{x}_i\right\|)$ are the cosine and Euclidean kernels respectively with corresponding fusion weights $\alpha_{1}$ and $\alpha_{2}$.

	Harnessing the distinctive properties of the cosine and Euclidean kernels, the formulation in (\ref{novel}) has shown some good results compared to the conventional Euclidean kernel (\cite{drmoin}).  We however argue that the fusion of the two kernels is manual and the weights $\alpha_{1}$ and $\alpha_{2}$ are adjusted in a hit-and-trial manner.  Without any prior information, a common practice is to assign equal weights to the two kernels i.e. $\alpha_1=\alpha_2=0.5$ .  As such, there is no dynamic method of optimizing these weights for a given data set.  We therefore propose a novel framework to adaptively optimize the weight assignment using the steepest descent method (\cite{thirtytwo}).
	
	The rest of the paper is organized as follows:  In Section \ref{sec:1} the proposed novel adaptive kernel is thoroughly discussed.  This is followed by extensive experiments in Section \ref{sec:2}.  The paper is finally concluded in Section \ref{conclusion}.

	\section{Proposed Method}
	\label{sec:1}
	
	We consider $\alpha_1$ and $\alpha_2$ in (\ref{novel}) to be dynamically adaptive variables:
	
	\begin{eqnarray}
	\label{1novel2}
	\alpha_1\equiv\frac{|\alpha_1(n)|}{|\alpha_1(n)|+|\alpha_2(n)|}
	\end{eqnarray}
	
	\begin{eqnarray}
	\label{2novel2}
	\alpha_2\equiv\frac{|\alpha_2(n)|}{|\alpha_1(n)|+|\alpha_2(n)|}
	\end{eqnarray}
	
	where the normalization of the mixing weights ensures that $\alpha_1(n)+\alpha_2(n)=1$.  The new kernel is therefore defined as:
	
	\begin{eqnarray}
	\label{novel2}
	\phi_i(\mathbf{x},\mathbf{x}_i)=\frac{|\alpha_1(n)|\phi_{i1}(\mathbf{x}.\mathbf{x}_i)+|\alpha_2(n)|\phi_{i2}(\left\|\mathbf{x}-\mathbf{x}_i\right\|)}{|\alpha_1(n)|+|\alpha_2(n)|}
	\end{eqnarray}
	
	The overall mapping, at the $n^{th}$ learning iteration linked to a specific epoch, can now be written as:
	\begin{eqnarray}
	\label{map}
	y(n)=\sum_{i=1}^{m_1}w_i(n)\phi_i(\mathbf{x},\mathbf{x}_i)+b(n)
	\end{eqnarray} 
	
	where the synaptic weights $w_i(n)$ and bias $b(n)$ are adapted at each iteration.  We define a cost function $\mathcal{E}(n)$ as:
	
	\begin{eqnarray}
	\label{cost}
	\mathcal{E}(n)=\mathcal{E}\left(\alpha_1(n),\alpha_2(n)\right)=\frac{1}{2}(d(n)-y(n))^{2}
	\end{eqnarray} 
	
	where $d(n)$ is the desired output at the $n^th$ iteration and $e(n)$ the instantaneous error between the desired output and the actual output of the neuron $e(n)=d(n)-y(n)$.  The update rule for the kernel's weight is given by: 
	
	\begin{eqnarray}
	\label{delta1}
	\Delta\alpha_1(n)=-\eta\frac{\partial \mathcal{E}(n)}{\partial \alpha_1(n)}
	\end{eqnarray}
	
	Using the chain rule of differentiation for the cost function in (\ref{cost}) yields:
	
	\begin{eqnarray}
	\label{partial1}
	\frac{\partial \mathcal{E}(n)}{\partial \alpha_1(n)}=\frac{\partial \mathcal{E}(n)}{\partial e(n)}\frac{\partial e(n)}{\partial y(n)}\frac{\partial y(n)}{\partial \phi_i(\mathbf{x},\mathbf{x}_i)}\frac{\partial \phi_i(\mathbf{x},\mathbf{x}_i)}{\partial \alpha_1(n)}
	\end{eqnarray}
	
	which upon the simplification of the partial derivatives in (\ref{partial1}) results in:
	
	\begin{multline}
	\label{partial2}
	\frac{\partial \mathcal{E}(n)}{\partial \alpha_1(n)}=-e(n)\Sigma_{i=1}^{m_i}w_i(n)\frac{|\alpha_1(n)||\alpha_2(n)|}{\alpha_1(n)[|\alpha_1(n)|+|\alpha_2(n)|]^{2}}\\ [\phi_{i1}(\mathbf{x}.\mathbf{x}_i)-\phi_{i2}(\left\|\mathbf{x}-\mathbf{x}_i\right\|)]
	\end{multline}
	
	and using (\ref{delta1}) and (\ref{partial2}) the update rule for $\alpha_1(n)$ is found to be:
	
	\begin{multline}
	\label{update1}
	\alpha_1(n+1)=\alpha_1(n) + \\ \eta e(n)\Sigma_{i=1}^{m_i}w_i(n)\frac{|\alpha_1(n)||\alpha_2(n)|}{\alpha_1(n)[|\alpha_1(n)|+|\alpha_2(n)|]^{2}}\\ [\phi_{i1}(\mathbf{x}.\mathbf{x}_i)-\phi_{i2}(\left\|\mathbf{x}-\mathbf{x}_i\right\|)]
	\end{multline} 
	
	Similarly the update rule for $\alpha_2(n)$ can be shown to be:
	
	\begin{multline}
	\label{update2}
	\alpha_2(n+1)=\alpha_2(n) + \\ \eta e(n)\Sigma_{i=1}^{m_i}w_i(n)\frac{|\alpha_1(n)||\alpha_2(n)|}{\alpha_2(n)[|\alpha_1(n)|+|\alpha_2(n)|]^{2}}\\ [\phi_{i2}(\left\|\mathbf{x}-\mathbf{x}_i\right\|)-\phi_{i1}(\mathbf{x}.\mathbf{x}_i)]
	\end{multline}
	
	The update equations of the weight and bias are given as:
	
	\begin{eqnarray}
	\label{upwt}
	w_i(n+1)=w_i(n)+\eta e(n)\phi_i(\mathbf{x},\mathbf{x}_i)
	\end{eqnarray}
	\begin{eqnarray}
	\label{bup}
	b_i(n+1)=b_i(n)+\eta e(n)
	\end{eqnarray}
	
	%
	The proposed approach is dynamic and does not require prior assignment of the weights for the participating kernels.

	\section{Experimental Results}
	\label{sec:2}
	The proposed novel kernel for the RBF is evaluated for three important tasks: (1) nonlinear system identification, (2) pattern recognition, and (3) function approximation.  All the experiments were conducted using Matlab on an Intel(R) Core(TM) i7-3770 CPU @ 3.4GHz machine with 4GB memory.
	
	\subsection{Nonlinear System Identification}
	
	\begin{figure}[h!]
		\begin{center}
			\centering 
			\includegraphics*[scale=0.5,bb=50 0 550 270]{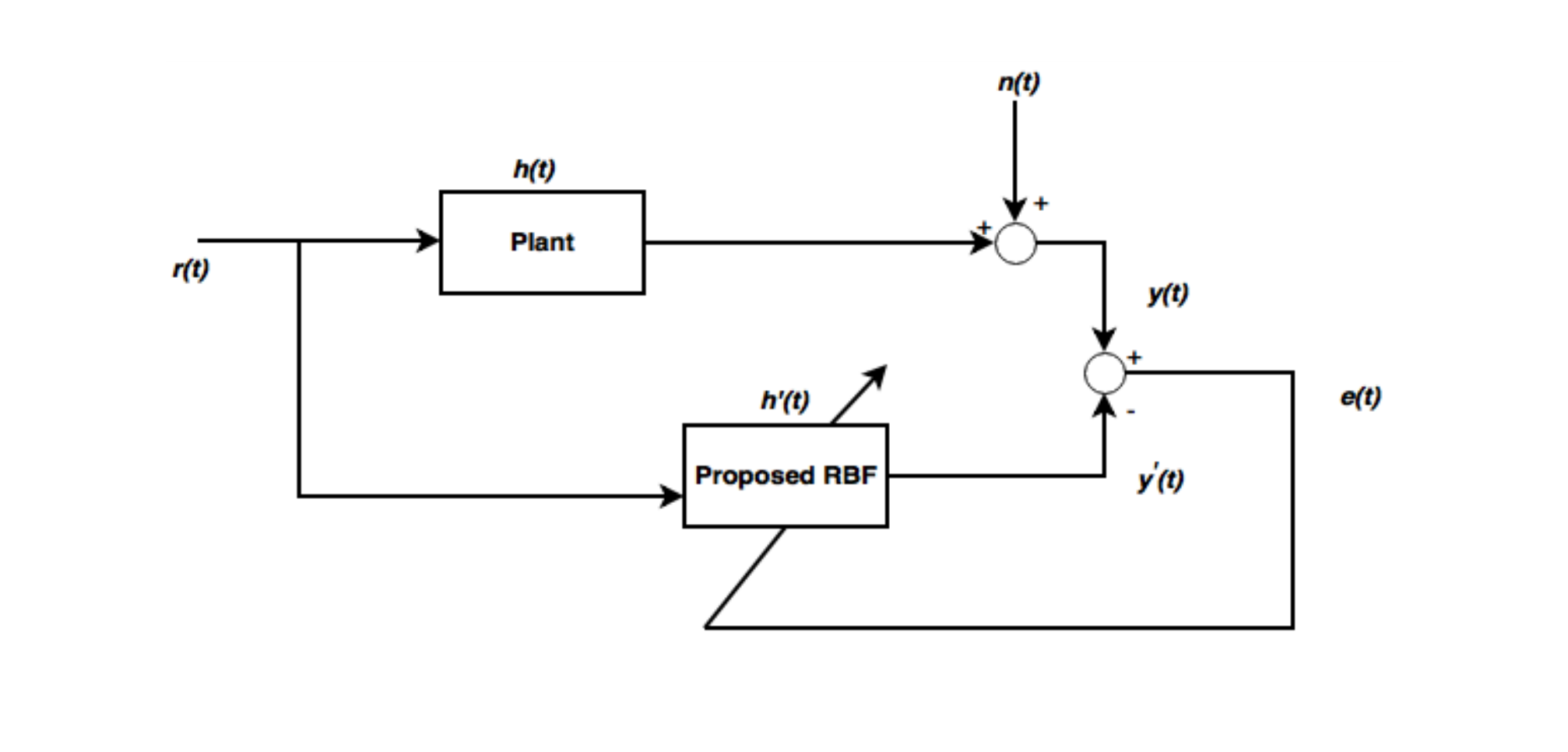} 
		\end{center}
		\caption{Nonlinear system identification using RBF neural network.}
		\label{plant1}
	\end{figure}
	Complex control systems and industrial processes can be effectively modeled using nonlinear systems (\cite{thirtysix}).  Nonlinear system identification is a method for estimating the mathematical model of a nonlinear system using the inputs and outputs to the system. RBF neural networks have shown to achieve good performance in this context (\cite{thirtyseven,thirtyeight,thirtynine}).  To evaluate the efficacy of the proposed novel kernel, we consider a highly nonlinear system, shown in Figure \ref{plant1}:
	
	\begin{multline}
	y(t)=a_1r(t)+a_2r(t-1)+a_3r(t-2)\\+a_4[\cos(br(t))+\exp(-|r(t)|)]+n(t)
	\end{multline}  
	
	where $r(t)$ and $y(t)$ are the input and output of the system respectively, $n(t)$ is the disturbance modeled assumed to be $\mathcal{N}(0,\sigma_d^2)$, $a_i$s are the polynomial coefficients describing the zeros of the system and $b>0$ is a constant.  For the purpose of this experiment $r(t)$ is taken to be a step function.  In Figure \ref{plant1}, the system is defined by its impulse response $h(t)$ while  $\hat{y}(t), \hat{h}(t)$ and $e(t)$ are the estimated output, estimated impulse response and the error of estimation respectively.  The simulation parameters chosen for the experiments are:  $a_1$=2, $a_2$=-0.5, $a_3$=-0.1, $a_4$=-0.7, $b=3$ and $\sigma_d^2$=0.0025.
	
	\begin{figure}[h!]
		\begin{center}
			\centering 
			\includegraphics*[scale=0.6,bb=20 0 450 340]{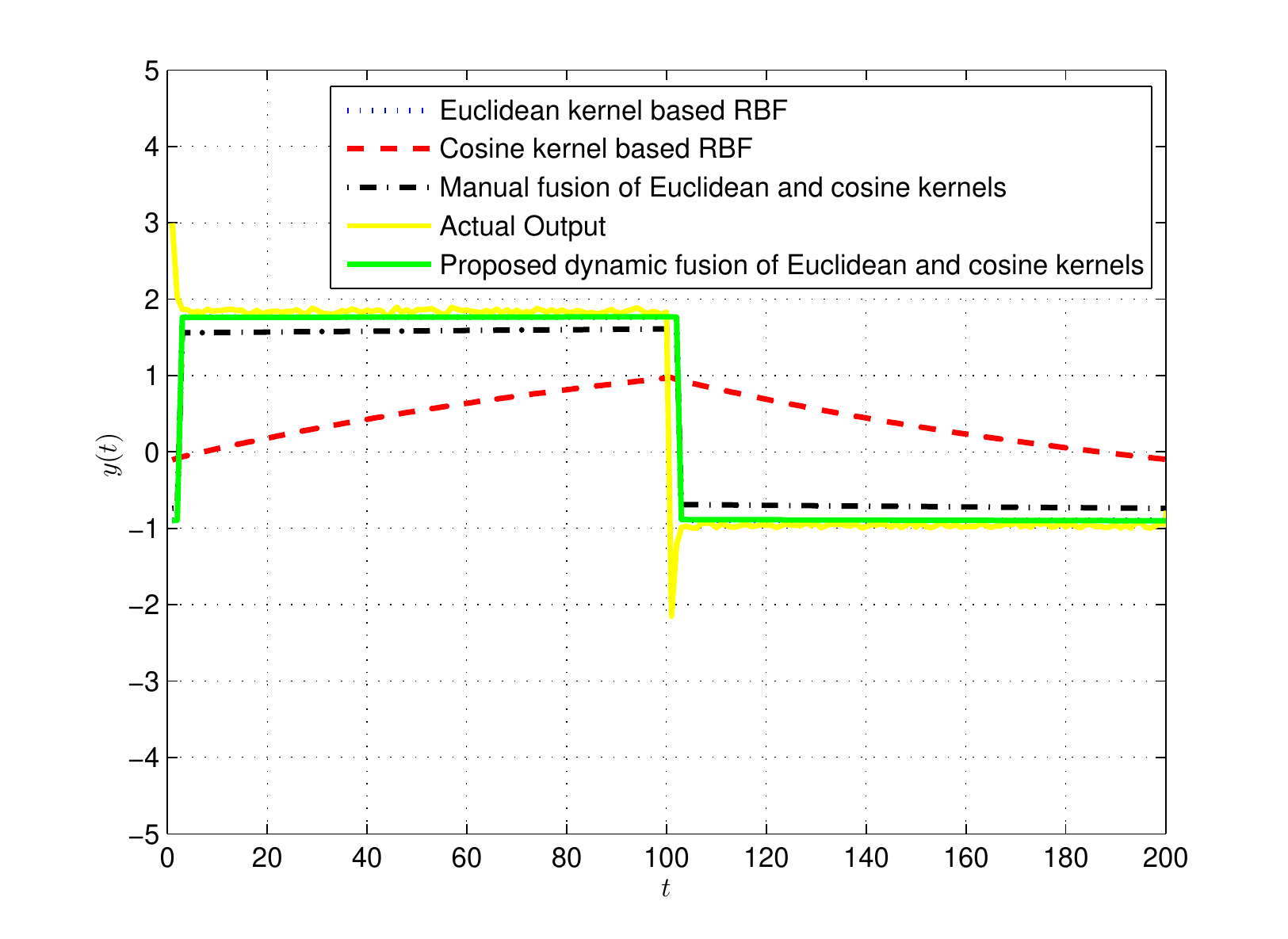} 
		\end{center}
		\caption{Comparison of the output of the nonlinear system.}
		\label{plantoutput}
	\end{figure}
	
	\begin{figure}[h!]
		\begin{center}
			\centering 
			\includegraphics*[scale=0.47,bb=0 0 500 400]{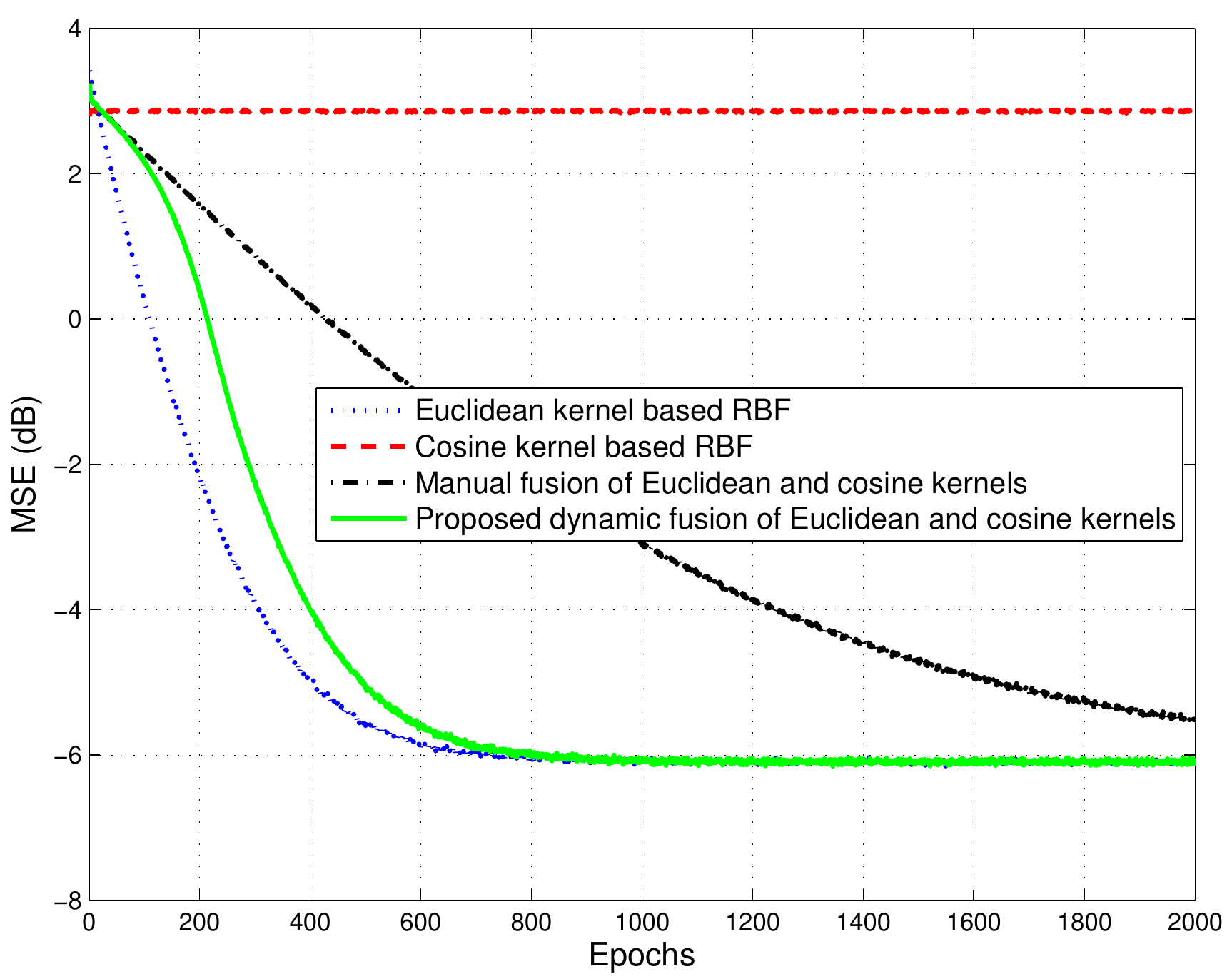} 
		\end{center}
		\caption{Nonlinear system:  The MSE curves for various approaches.}
		\label{planterror}
	\end{figure}
	
	\begin{figure}[h!]
		\begin{center}
			\centering
			\includegraphics*[scale=0.47,bb=0 0 500 400]{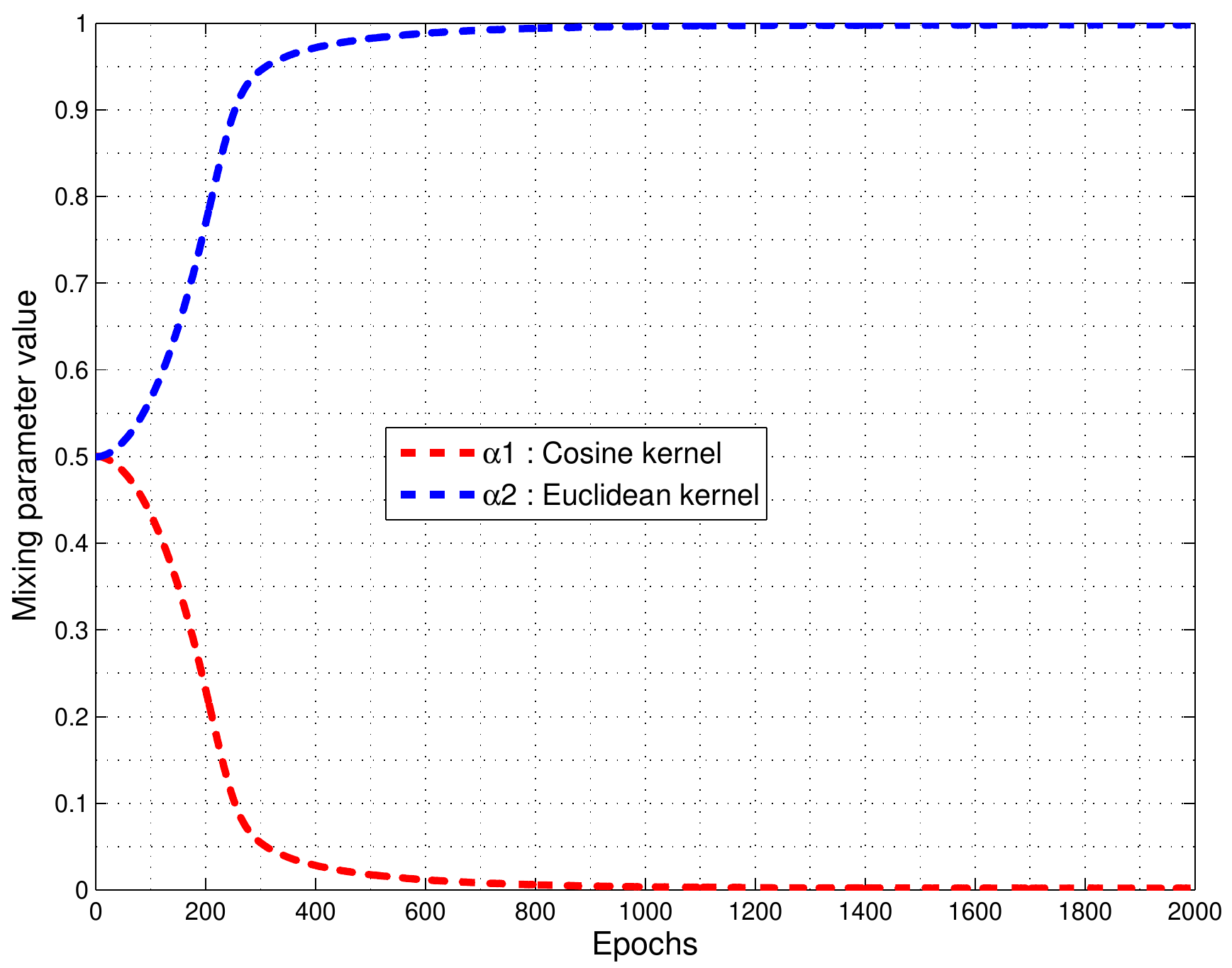}
		\end{center}
		\caption{Nonlinear system:  Adaptation of the mixing parameters with respect to time.}
		\label{plantALPHA}
	\end{figure}
	
	For the RBF structure, the number of neurons were selected to be 401 and the centers were uniformly spaced between -50 to 50 with a step size of 0.25. The initial weights and bias values were initialised to zero. For the Gaussian kernel the spread $\sigma$ was set to 0.1 and for the cosine kernel a small value of $\gamma= 1e-50$ was used.  For the proposed approach, the initial values of $\alpha_1$ and $\alpha_2$ are taken to be 0.5. 
	Figure \ref{plantoutput} shows the estimated output of the proposed approach compared to the actual output, the Euclidean kernel ($\alpha_1=0, \alpha_2=1$), cosine kernel ($\alpha_1=1, \alpha_2=0$) and the manual fusion of the two kernels ($\alpha_1=\alpha_2=0.5$).  Note that due to the most precise estimation, the Euclidean kernel overlaps the actual output and therefore cannot be distinguished.  The mean square error (MSE) curves are depicted in Figure \ref{planterror}.  The Euclidean kernel produces the best performance achieving a minimum MSE of -6.1943 dB in 1379 iteration epochs, while the cosine kernel performs poorly with a MSE of 2.7887 dB.  Without any prior information, the proposed approach dynamically gives more weight to the Euclidean kernel, attaining a minimum mean square error (MSE) of -6.1547 dB in 1447 iterations which is quite comparable to the Euclidean kernel. The final values of the weights were found to be $\alpha_1=0.002$ and $\alpha_2=0.998$.  The proposed approach is substantially better compared to the manual fusion of kernels which achieved a minimum mean square error (MSE) of -5.5176 dB in 1992 iterations.  Variation of the mixing parameters with respect to the iteration epochs is depicted in Figure \ref{plantALPHA}.  For the comparison of time complexity of the proposed method with manual fusion of the two kernels, we investigated the training time for 2000 epochs.  The proposed method utilizes 550.78 seconds whereas the manual fusion of the two kernel takes 537.74 seconds.  The experiment clearly shows that in the absence of any prior knowledge, the proposed approach adaptively emphasizes the effective Euclidean kernel and achieves a comparable performance.   
	

	\subsection{Pattern Classification}
	
\begin{figure}[h!]
		\begin{center}
			\centering
			\includegraphics*[scale=0.47,bb=0 0 500 400]{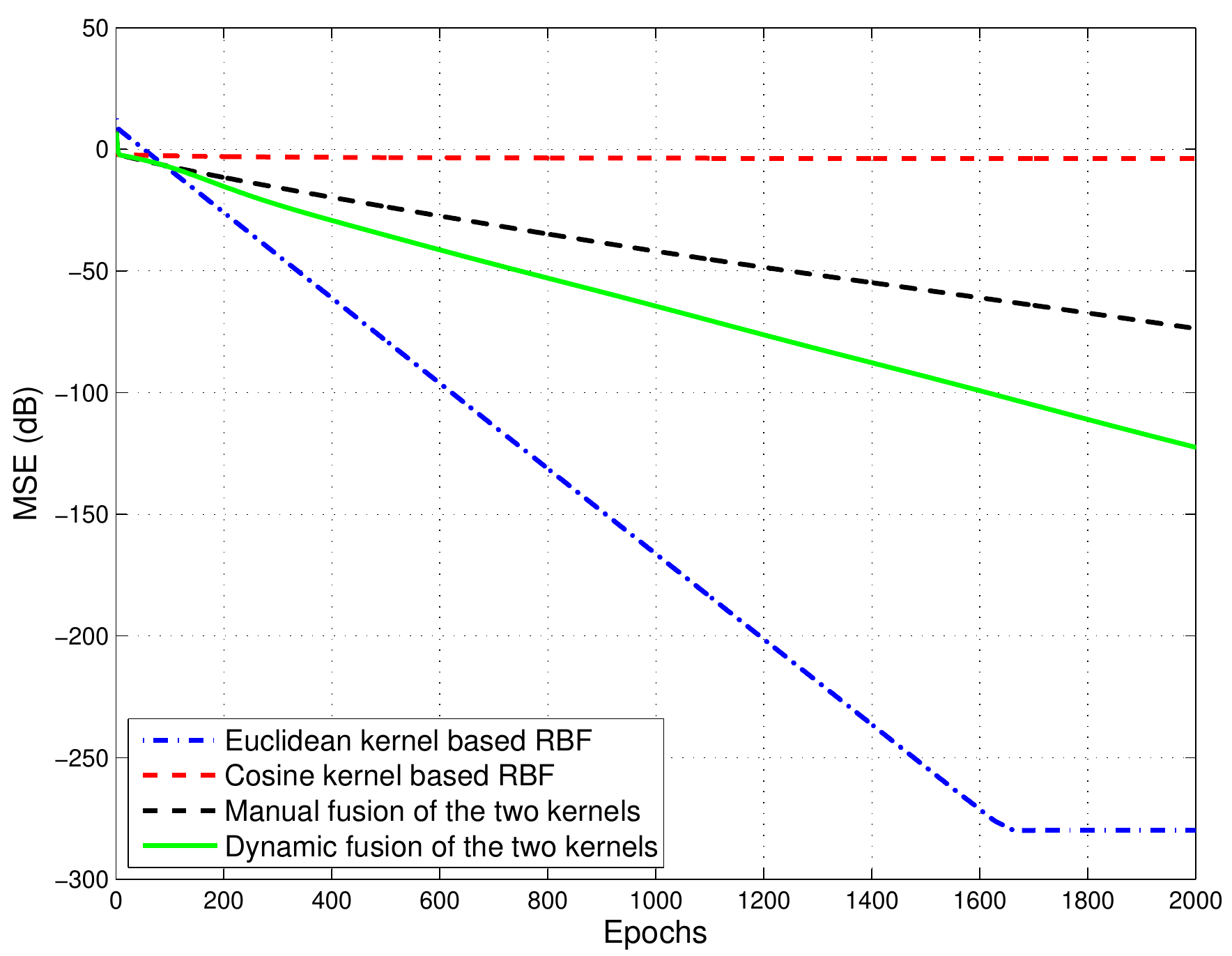}
		\end{center}
		\caption{The MSE curves for the training phase of the pattern classification problem.}
		\label{error_1}
	\end{figure}

	Machine learning methods have been used with great success in bioinformatics (\cite{peng}).  One of the important applications is the prediction of cancer using gene micro array data.  In this experiment we target the prediction of leukemia disease using the standard  Leukemia ALL/AML data (\cite{leukemia}).  The data set consists of 38 training samples from bone marrow specimens (27 ALL and 11 AML) and 34 testing samples.  There are 34 test samples (20 ALL and 14 AML) prepared under different experimental conditions including 24 bone marrow and 10 blood sample specimens.  The data set consists of 7129 genes. The Minimum Redundancy and Maximum Relevance (mRMR) is an established technique to select the most significant genes (\cite{peng}).  The mRMR technique was used to select only the top five genes for our experiments.
	For the RBF structure, the number of neurons were selected to be 38 and the centers were chosen using the subtractive clustering method of \cite{sub_clust} with an influence factor of 0.1. The initial weights and bias values were initialised to zero. For the Gaussian kernel the spread $\sigma$ was set to 0.2 and for the cosine kernel a small value of $\gamma= 1e-50$ was used. For the proposed approach, the initial values of $\alpha_1$ and $\alpha_2$ are taken to be 0.5. 
	For the training phase, the MSE curves of different approaches are shown in Figure \ref{error_1}.
	
	\begin{figure}[h!]
		\begin{center}
			\centering
			\includegraphics*[scale=0.47,bb=0 0 500 400]{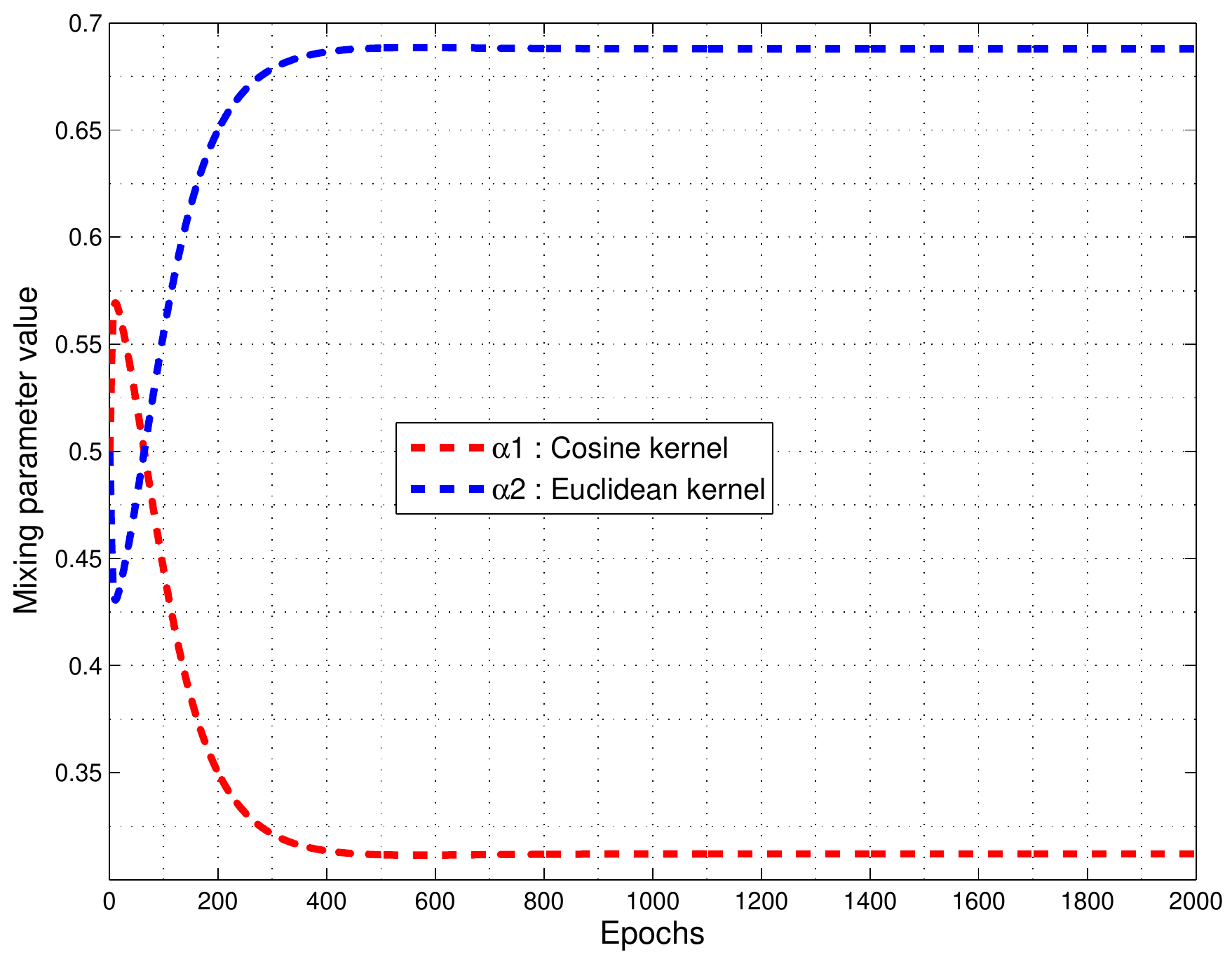}
		\end{center}
		\caption{Pattern classification: Adaptation of the mixing parameters with respect to time.}
		\label{ClassificationALPHA}
	\end{figure}
	
	The Euclidean kernel outperforms the cosine kernel achieving a minimum MSE of -279.9331 dB. The proposed method dynamically gives more weight to the Euclidean kernel achieving an MSE of -122.4990 dB with $\alpha_1=0.3121$ and $\alpha_2=0.6879$.  Note that although the Euclidean kernel achieves the minimum MSE for the training data, it is merely the case of overfitting where a classifier achieves the best performance on the training set but fails on the test data.  Variation of the mixing parameters with respect to epochs is depicted in Figure \ref{ClassificationALPHA}.  In Figure \ref{error_1}, after the $65^{th}$ epoch the MSE of Euclidean kernel becomes lower than the cosine kernel, noteworthy is the corresponding flip in the weights adaptively assigned by the proposed approach in Figure \ref{ClassificationALPHA}.  Note that the weights become stable after 400 epochs.  The manual fusion of the two kernels ($\alpha_1=\alpha_2=0.5$) results in an MSE of -73.6652 dB which is inferior to the proposed method.  The training accuracies of all the approaches are presented in Table \ref{pat1}, note that all approaches result in 100\% accuracy for the training samples.  The total training time for the proposed method is found to be 12.98 seconds whereas the manual fusion of the two kernel takes 12.65 seconds.
	
	\begin{table*}[h!]
		\centering
		\caption{Results for the pattern classification problem.}  
		\label{pat1} 
		\begin{tabular}{c|c|c}
			\hline
			{\bf Approach} & {\bf Training Accuracy} & {\bf Testing Accuracy} \\
			\hline
			{ Cosine kernel} &   100.00\% &    94.12\%  \\
			\hline
			{ Euclidean kernel} &   100.00\% &  58.82\%  \\
			\hline
			{ Manual fusion of the two kernels} &   100.00\% &    94.12\% \\
			\hline
			{\bf Proposed dynamic fusion} &   \bf 100.00\% &    \bf 97.06\% \\
			\hline
		\end{tabular}

	\end{table*}
	
	True evaluation of any predictive system is for the case of unseen samples i.e the ``testing phase".  Although the Euclidean kernel achieves the minimum MSE during the training phase, the proposed approach demonstrated that the best performance for the testing stage is achieved with an accuracy of 97.06\%.  The Euclidean kernel was trained ``too well" on the training samples and therefore incurred the problem of ``overfitting" attaining a test accuracy of only 58.82\%.  The proposed dynamic fusion of the two kernels outperformed the manual fusion ($\alpha_1=\alpha_2=0.5$) by a margin of 2.94\%.   
	
	We provide an intuitive understanding of the proposed approach using this pattern classification problem.  The data which is not linearly separable in the original space poses a challenging task in the classification theory.  Cover's theorem states that such data can be mapped into a high dimensional space using a nonlinear mapping function (kernel function), thereby resulting in a linearly separable data in the transformed space. 
	
\begin{figure}[h!]
		\begin{center}
			\centering 
			\includegraphics*[scale=0.5,bb=0 0 500 400]{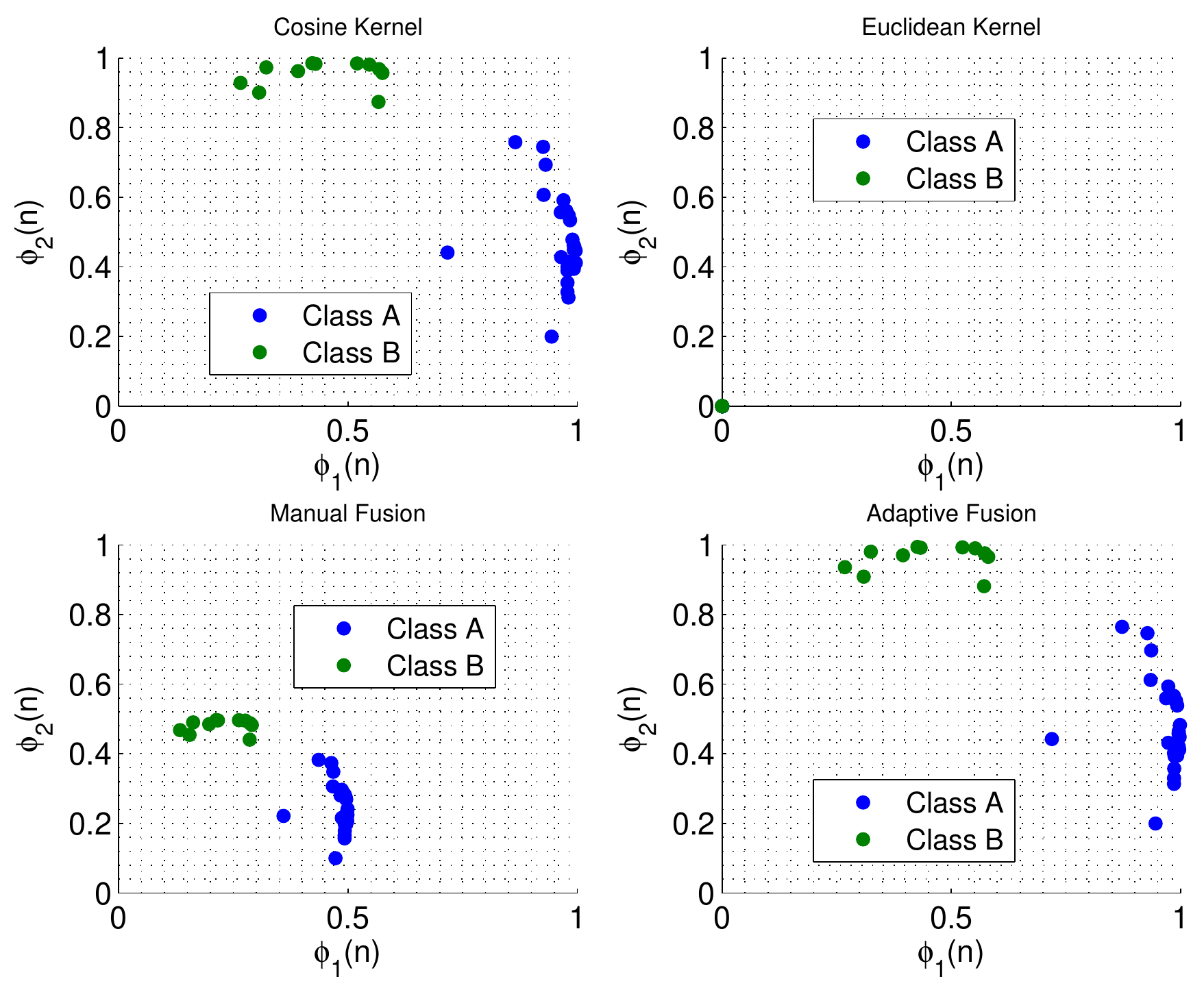} 
		\end{center}
		\caption{Clustering of the Leukemia data using various kernels.  Note the class separation for the proposed kernel}
		\label{clustering}
	\end{figure}

	Selection of an appropriate kernel is an important issue to be considered.  A good kernel will result in optimal separation of classes in the transformed space thereby improving the performance on unseen test samples.  Using fusion of multiple kernels is often a good idea to harness the complementary properties of various kernels.  The weights of the combining kernels play an important role in such cases.  Selecting weights on random bases may result in an inefficient fusion.   The proposed adaptive fusion framework automatically selects the best weights for the combining kernels resulting in maximum separation of classes.  
	We demonstrate this through clustering of the Leukemia dataset consisting of 38 samples (27 Class A and 11 Class B) and 5 attributes. For demonstration purposes we choose two centers $c_1$ and $c_2$ which are the means of classes A and B respectively.  The mapping of the samples in the 2D - space using various kernels is shown in Figure \ref{clustering}.
	
	It can be seen that cosine kernel efficiently separates the two classes in the 2D-space while the Euclidean kernel maps all the samples to origin (overlapping samples seen as one green circle).  The manual fusion of the kernels (with equal weights) results in a decreased class separation compared to the cosine kernel.  The proposed adaptive fusion of the two kernels automatically assigns more weight to the cosine kernel thereby resulting in better clustering compared to the manual fusion. 
	
	\subsection{Function Approximation}
	
	We consider the problem of approximation of a non-linear function defined by:
	
	\begin{eqnarray}
	\label{RegFUN}
	f(x,y)=\exp(x^2-y)
	\end{eqnarray}

\begin{figure}[h!]
		\begin{center}
			\centering
			\includegraphics*[scale=0.47,bb=0 0 500 400]{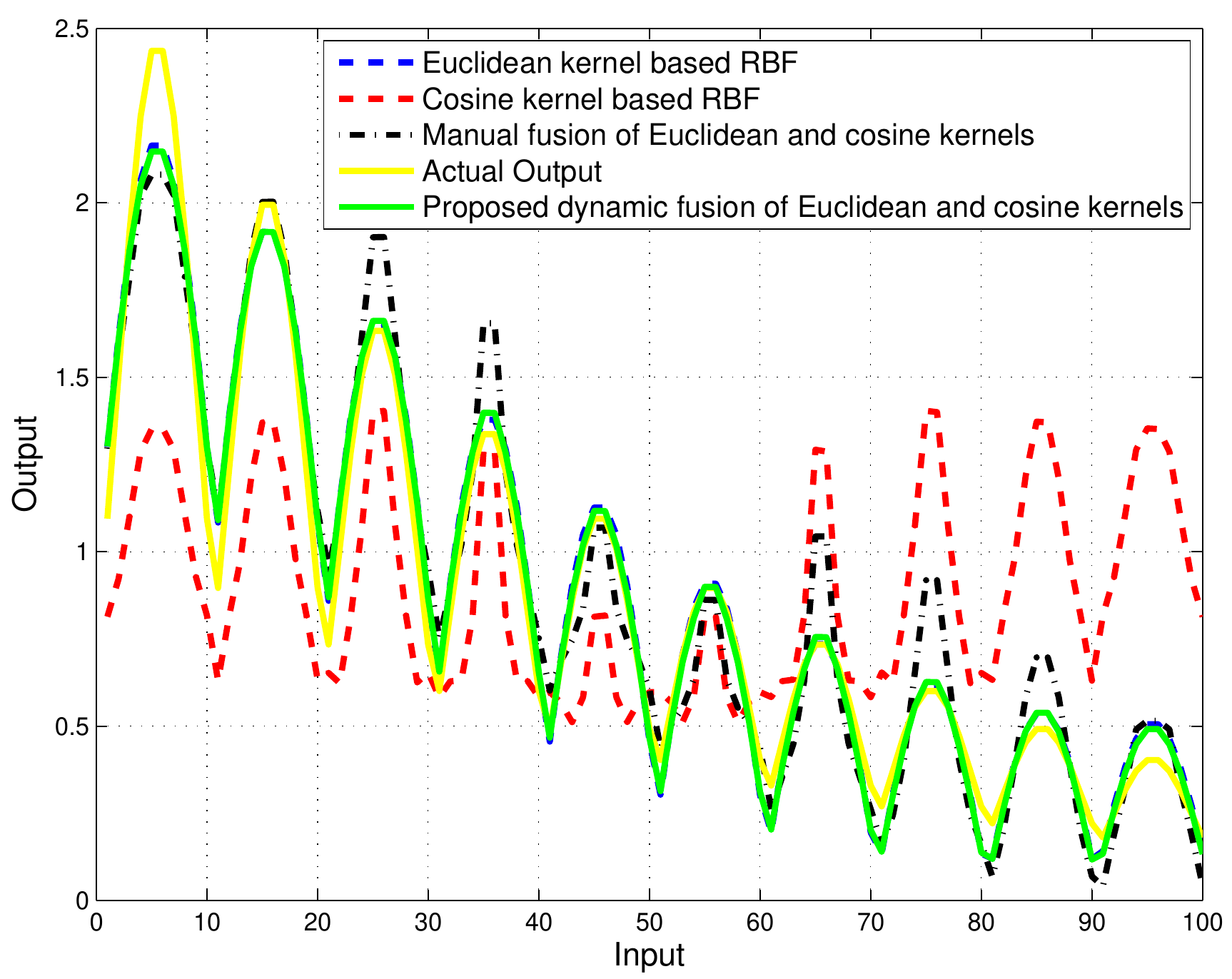}
		\end{center}
		\caption{Comparison of the output of the nonlinear function for various kernels.}
		\label{RegOUT}
	\end{figure}

	The function in equation (\ref{RegFUN}) is approximated using various kernels.  
	For all experiments 121 centers were considered and the learning rate was taken to be $\eta=1\times 10^{-3}$.  The centers were chosen through the subtractive clustering method of \cite{sub_clust} with an influence factor of 0.1.  The initial weights and bias values were initialised to zero. For the Gaussian kernel the spread $\sigma$ was set to 0.2 and for the cosine kernel a small value of $\gamma= 1e-50$ was used.  For the proposed approach, the initial values of $\alpha_1$ and $\alpha_2$ are taken to be 0.5. 
	A total of 121 values of $x$ and $y$ are used for training ranging from -1 to 1 with a step size of 0.2.  Testing has been conducted on 100 data points ranging from -0.9 to 0.9 with a step size of 0.2.  For the test data, Figure \ref{RegOUT} shows the estimated output of the proposed approach compared to the actual output, the Euclidean kernel ($\alpha_1=0, \alpha_2=1$), the cosine kernel ($\alpha_1=1, \alpha_2=0$) and the manual fusion of the two kernels ($\alpha_1=\alpha_2=0.5$) in reduced dimension.  
	
	\begin{figure}[h!]
		\begin{center}
			\centering
			\includegraphics*[scale=0.47,bb=0 0 500 400]{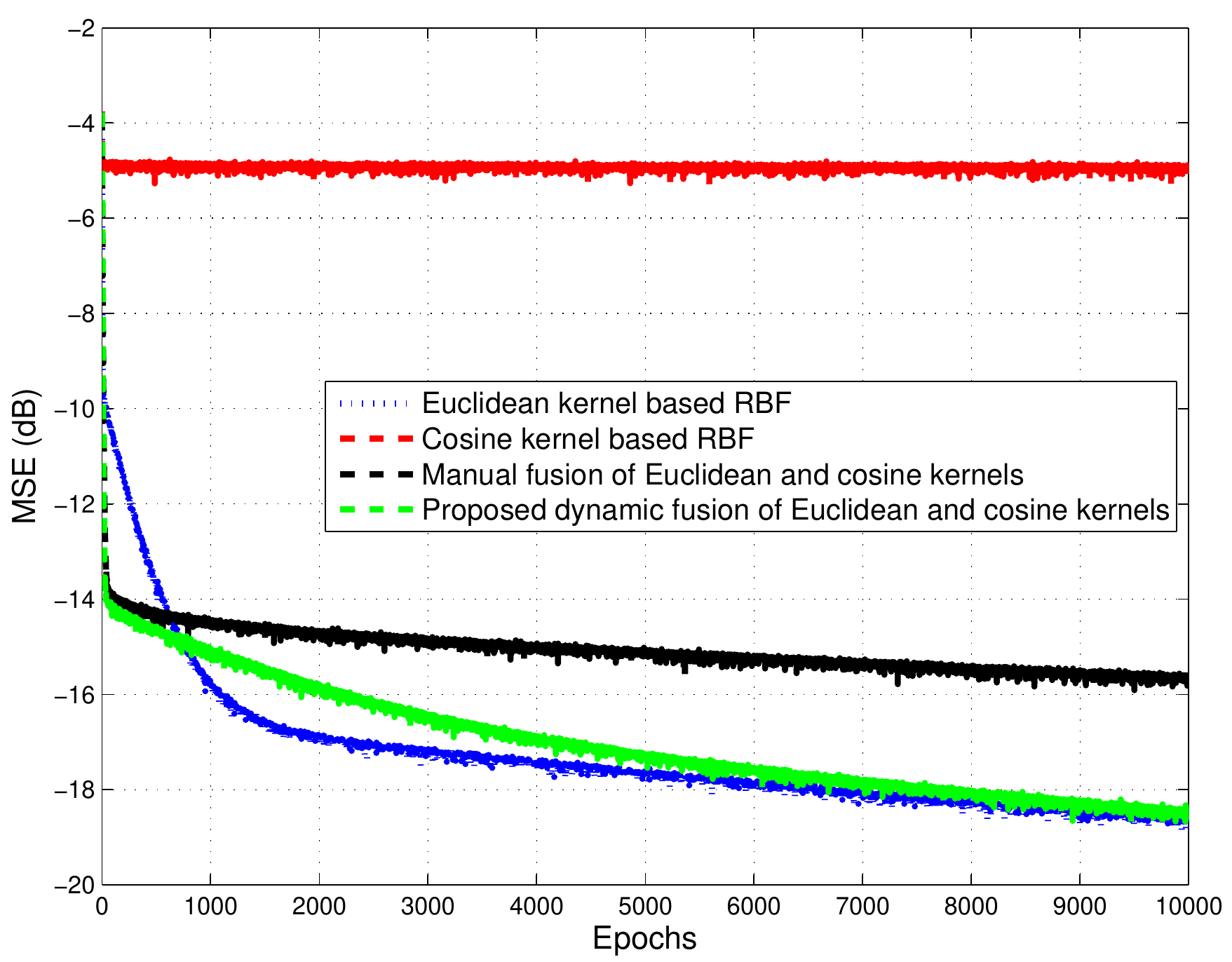}
		\end{center}
		\caption{The MSE curves for the training phase of the function approximation problem.}
		\label{RegMSE}
	\end{figure}
	
	\begin{figure}[h!]
		\begin{center}
			\centering
			\includegraphics*[scale=0.47,bb=0 0 500 400]{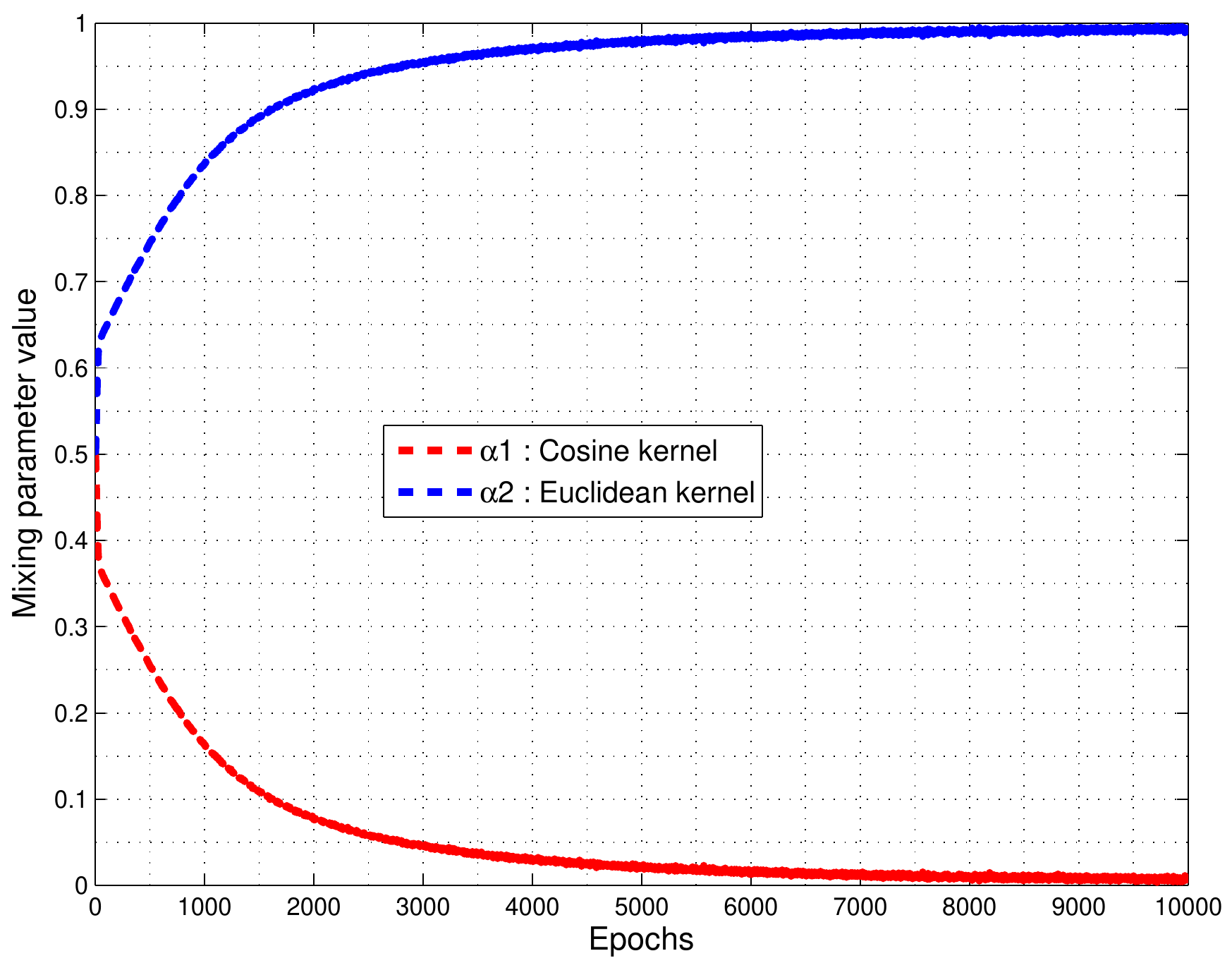}
		\end{center}
		\caption{Function approximation: Adaptation of the mixing parameters with respect to time.}
		\label{RegALPHA}
	\end{figure}
	
	The mean square error (MSE) curves are depicted in Figure \ref{RegMSE}.  The Euclidean kernel produces the best performance achieving a minimum MSE of -18.6619 dB, while the cosine kernel performs poorly with an MSE of -4.9277 dB.  Without any prior information, the proposed approach dynamically gives more weight to the Euclidean kernel. The proposed approach attains a minimum mean square error (MSE) of -18.4076 dB which is comparable to the Euclidean method.  The proposed approach is substantially better compared to the manual fusion of kernels which achieved a minimum mean square error (MSE) of -15.6181 dB.  Variation of the mixing parameters with respect to the iteration epochs is depicted in Figure \ref{RegALPHA}.  The final values of the weights were found to be $\alpha_1=0.0060$ and $\alpha_2=0.9940$. The experiment clearly shows that in the absence of any prior knowledge, the proposed approach adaptively emphasizes the effective Euclidean kernel and achieves better performance.  For the comparison of the time complexity of the proposed method with manual fusion of the two kernels, we investigated the training time for 10000 epochs.  The total training time for the proposed method is found to be 586.3 seconds whereas the manual fusion of the two kernel takes 578.2 seconds.

	\section{Conclusion}\label{conclusion}
	In this research a novel kernel for the RBF neural network is proposed.  The proposed framework adaptively fuses the Euclidean and cosine distance measures thereby harnessing the complementary properties of the two.  The proposed algorithm is dynamic and adaptively learns the optimum weights of the participating kernels for a given problem.  The efficacy of the proposed kernel is demonstrated on three important problems, namely nonlinear system identification, pattern classification and function approximation.  The proposed algorithm has shown to comprehensively outperform the manual fusion of the two kernels.  For the problem of nonlinear system identification, the proposed framework adaptively assigns a higher fusion weight to the Euclidean kernel achieving a comparable performance. The proposed algorithm performs better than the manual fusion of the two kernels.  Therefore, in the absence of any prior knowledge, the proposed method is shown to emphasize the most effective kernel.  For the pattern classification problem, the proposed method dynamically assigns more weight to the Euclidean kernel and achieves a comparable training accuracy of 100\%.  For the more challenging testing phase, the proposed optimized fusion attains the best accuracy of 97.06\%.  Note that the proposed approach outperformed the best conventional kernel i.e. the Euclidean kernel by meaningfully utilizing the complementary properties of the cosine kernel.  For the function approximation problem, the Euclidean kernel produces the best performance achieving a minimum MSE of -18.6619 dB, while the cosine kernel performs poorly with an MSE of -4.9277 dB. Without any prior information, the proposed approach dynamically gives more weight to the Euclidean kernel and achieved a minimum MSE of -18.4076 dB.  The experiments clearly demonstrate that the proposed optimum fusion of kernels will always perform equal to or better than the best participating kernel. 
	\section{Acknowledgement}
	
	The authors would like to thank University of Western Australia (UWA), Pakistan Air Force - Karachi Institute of Economics and Technology (PAF-KIET), and Iqra University (IU), for providing the necessary support towards conducting this research and the anonymous reviewers for their important comments.
	
	\bibliographystyle{IEEEtran}
	\bibliography{IEEE_FORMAT_ARXIV}

\end{document}